\documentclass[letterpaper]{article} 
\usepackage{aaai25}  
\usepackage{times}  
\usepackage{helvet}  
\usepackage{courier}  
\usepackage[hyphens]{url}  
\usepackage{graphicx} 
\usepackage{natbib}  
\usepackage{caption} 
\usepackage{algorithm}
\usepackage{algorithmic}
 
\usepackage{amsmath,amssymb}

\usepackage{newfloat}
\usepackage{listings}
\DeclareCaptionStyle{ruled}{labelfont=normalfont,labelsep=colon,strut=off} 
\floatstyle{ruled}
\newfloat{listing}{tb}{lst}{}
\floatname{listing}{Listing}
\title{Learning to Manipulate Under Limited Information}
\author {
    Wesley H. Holliday\textsuperscript{\rm 1},
    Alexander Kristoffersen\textsuperscript{\rm 1},
    Eric Pacuit\textsuperscript{\rm 2}
}
\affiliations {
    \textsuperscript{\rm 1}University of California, Berkeley\\
    \textsuperscript{\rm 2}University of Maryland\\
    wesholliday@berkeley.edu, akristoffersen@berkeley.edu, epacuit@umd.edu
}

\usepackage{bibentry}

\begin{document}

\maketitle

\begin{abstract}
By classic results in social choice theory, any reasonable preferential voting method sometimes gives individuals an incentive to report an insincere preference. The extent to which different voting methods are more or less resistant to such strategic manipulation has become a key consideration for comparing voting methods. Here we measure resistance to manipulation by whether neural networks of various sizes can learn to profitably manipulate a given voting method in expectation, given different types of limited information about how other voters will vote. We trained over 100,000 neural networks of 26 sizes to manipulate against 8 different voting methods, under 6 types of limited information, in committee-sized elections with 5--21 voters and 3--6 candidates. We find that some voting methods, such as Borda, are highly manipulable by networks with limited information, while others, such as Instant Runoff, are not, despite being quite profitably manipulated by an ideal manipulator with full information. For the three probability models for elections that we use, the overall least manipulable of the 8 methods we study are Condorcet methods, namely Minimax and Split~Cycle.\end{abstract}

%
\begin{links}
     \link{Code}{https://github.com/epacuit/ltm}
\end{links}

\section{Introduction}

A fundamental problem in multi-agent decision making is that of aggregating heterogeneous preferences  \cite{Conitzer2010}. Voting theory provides many possible methods of preference aggregation with different benefits and costs \cite{Zwicker2016}. However, no reasonable preferential voting method escapes the problem of \textit{manipulability}. As shown by classic results such as the Gibbard-Satterthwaite theorem \cite{Gibbard1973,Satterthwaite1973} and its generalizations \cite{duggan2000strategic,Taylor2005}, for any such voting method, there is some preference profile in which some voter has an incentive to report an insincere preference in order to obtain a result that is preferable, according to their sincere preference, to the result they would obtain if they were to submit their sincere preference. Thus, sincere voting is not a Nash equilibrium of the game derived from this preference profile, where the players are the voters and the actions are the possible preference rankings to report.

However, the mere existence of a preference profile in which a voter has an incentive to misreport their preferences tells us little about the frequency with which a voter will have such an incentive or the difficulty of recognizing that such an incentive exists---under either full information or limited information about the profile. Thus, a manipulable voting method might be relatively \textit{resistant} to manipulation, either because the frequency just cited is low, the difficulty just cited is high, or a mixture of both factors. Such resistance to manipulation has been considered an important criterion for comparing voting rules \cite{Merrill1988,greenarmytage2016statistical}.

As for the difficulty of manipulation, there is now a large literature in computational social choice on the worst-case  complexity of computing whether there is a strategic ranking that will elect a desired candidate.   Faliszewski and Procaccia \citeyearpar{faliszewski2010ai} call this ``AI's war on manipulation.'' A series of hardness results have been proved since \citealt{Bartholdi1989} and \citealt{conitzer2007elections}. There is evidence for and against the view that high worst-case computational complexity is a barrier to manipulation \cite{walsh2011computational}, and the situation with average-case manipulability might be quite different \cite{mossel2013}.

In this paper, we take a different approach than previous work, measuring resistance to manipulation by whether neural networks of various sizes can learn to profitably manipulate a given voting method in expectation, given different types of information about how other voters will vote. Like the classic results on manipulation, we focus on the case of a single manipulating voter. A single voter can almost never affect the outcome of a large election with thousands of voters or more by changing their vote, so rational manipulation by a single voter is most relevant for small elections in committees, boards, etc. The classic results on manipulation also in effect assume that the manipulator knows exactly how all other voters will vote, which is unrealistic in many voting contexts. By contrast, we train neural networks to manipulate on the basis of different types of limited information.

\subsection{Related work}

\subsubsection{Manipulation under Limited Information}

A number of previous papers study whether a voter can successfully manipulate in an election under limited information about how other voters will vote \cite{myerson1993theory,conitzer2011dominating,reijngoud2012voter,meir2014localdominance,endriss2016strategic,lang2020collective,Veselova2023}, including through the use of heuristics  \cite{chopra2004knowledge,laslier2009leader,meir2018strategic,fairstein2019modeling}. The information of the manipulator is typically represented by a set of preference profiles, all of which agree on (i) the manipulator's own preferences and (ii) some other partial information (e.g., all profiles in the manipulator's information set agree on who will win the election if the manipulator votes sincerely, or all profiles in the set are such that for each of the other voters $i$, $i$'s ranking in the profile extends some known partial order over the candidates that the manipulator attributes to $i$), perhaps supplemented with a probability measure over such profiles \cite{lu2012bayesian}. By contrast, in this paper we will represent limited information by the inputs to a neural network.

\subsubsection{Machine Learning and Voting Theory}

Several previous works apply machine learning to problems in voting theory, though not in the way we do here. \citealt{anil2021learning}, \citealt{burka2022voting}, and \citealt{Matone2024} study the learnability of various voting methods; \citealt{kang2023learning} studies learning how to explain election results for various voting methods; and \citealt{Armstrong2019} and \citealt{Firebanks2020} use machine learning to create new voting methods satisfying desiderata. But none of these papers discuss learning to manipulate as a voter.  Manipulation is studied in \citealt{Airiau2017}, but only in the context of \textit{iterative voting},\footnote{In iterative voting, after all voters submit rankings and a tentative winner is announced, the voters are allowed to sequentially change their rankings, with a new tentative winner announced after each change, until an equilibrium is reached.} whereas we focus on learning to manipulate in traditional elections, where the final winner is immediately computed after all voters submit their rankings.

\subsubsection{Machine Learnability as a Metric of Task Difficulty}

Sufficiently large neural networks are able to learn arbitrarily complex functions, including fitting to random data \cite{zhang2017understanding}. If the model is not large enough to fully memorize the training data, however, learning the training data requires generalization. In the fields of reinforcement learning and natural language processing, it is commonly held that more complex problems may require larger and more complex networks. Indeed, previous work has shown that model performance grows as the number of learnable parameters increases \cite{HestnessArticle,golubeva2021wider}.  In this paper, we use required model size as a proxy for task difficulty. Not only has learnability by neural networks been taken to be suggestive of human learnability \cite{Steinert-Threlkeld2020}, but also humans may use a neural network to help them manipulate an election under limited information.

\section{Preliminaries}

Given a set $V$ of voters and a set $X$ of candidates, a \textit{preference profile for $(V,X)$} is a function $\mathbf{P}$ assigning to each $i\in V$ a linear order $\mathbf{P}_i$ of $X$. Where $\mathbf{Q}$ is a preference profile for $(V,X)$, $i\not\in V$, and $\mathbf{P}_i$ is a linear order of $X$, we write $(\mathbf{P}_i, \mathbf{Q})$ for the preference profile that assigns to $i$ the linear order $\mathbf{P}_i$  and assigns to each $j\in V$ the linear order~$\mathbf{Q}_j$.   For a profile $\mathbf{P}$ and voter $i\in V$, let $\mathbf{P}_{-i}$ be the profile for $(V\setminus\{i\},X)$ obtained by restricting $\mathbf{P}$ to all voters except $i$, so we may write $\mathbf{P}=(\mathbf{P}_i,\mathbf{P}_{-i})$. For a candidate $a\in X$, let $\mathbf{P}_{-a}$ be the profile for $(V,X\setminus\{a\})$ obtained from $\mathbf{P}$ by restricting each voter's linear order to $X\setminus\{a\}$.

Given a profile $\mathbf{P}$ and candidates $a,b\in X$, the \textit{margin of $a$ vs.~$b$} in $\mathbf{P}$, denoted $\mathrm{Margin}_\mathbf{P}(a,b)$, is the number of voters who rank $a$ above $b$ minus the number who rank $b$ above $a$ in $\mathbf{P}$. A \textit{Condorcet winner} in $\mathbf{P}$ is a candidate $c\in X$ with a positive margin over every $a\in X\setminus\{c\}$.

A \textit{utility profile for $(V,X)$} is a function $\mathbf{U}$ assigning to each $i\in V$ a utility function $\mathbf{U}_i:X\to \mathbb{R}$, where we assume that ${\mathbf{U}_i(x)\neq \mathbf{U}_i(y)}$ whenever  $x\neq y$.\footnote{The probability models for utility profiles that we use make a tie in the utilities of distinct candidates a measure zero event.} Given such a utility profile $\mathbf{U}$, its \textit{induced preference profile $\mathbf{P}(\mathbf{U})$} assigns to each $i\in V$ the linear order $\succ_i$ defined by 
\begin{equation}\mbox{$x\succ_i y$ iff $\mathbf{U}_i(x)>\mathbf{U}_i(y)$.}\label{InducePref}\end{equation}

A (\textit{preferential}) \textit{voting method for $(V,X)$} is a function $F$ whose domain is the set of preference profiles for $(V,X)$ such that for any $\mathbf{P}\in\mathrm{dom}(F)$, we have $\varnothing\neq F(\mathbf{P})\subseteq X$. We list the voting methods we study in the next subsection.

In case $F(\mathbf{P})$ has more than one element, we assume an even-chance lottery $F_\ell(\mathbf{P})$ on $F(\mathbf{P})$ determines the ultimate tiebreak winner. Thus, given a utility function $\mathbf{U}_i$ on $X$, the expected utility of this lottery is given by
\[\mathbf{EU}_i(F_\ell(\mathbf{P}))= \frac{\sum_{a\in F(\mathbf{P})} \mathbf{U}_i(a)}{|F(\mathbf{P})|}.\]

Given a voting method $F$, utility profile $\mathbf{U}$ for $(V,X)$ with $\mathbf{P}=\mathbf{P}(\mathbf{U})$, and voter $i\in V$, we say that a linear order $\mathbf{P}_i'$ of $X$ is a \textit{profitable manipulation of $F$ at $\mathbf{U}$ by $i$} if 
\begin{equation}\mathbf{EU}_i (F_\ell(\mathbf{P}'_i, \mathbf{P}_{-i})) > \mathbf{EU}_i (F_\ell(\mathbf{P})).\label{ManipEq}\end{equation}
We say $\mathbf{P}_i'$ is \textit{optimal} if the left-hand side of (\ref{ManipEq}) is maximized for $\mathbf{P}'_i$ among all possible linear orders of $X$. We assume, as in standard decision theory \cite{Kreps1988}, that our manipulating agent aims to maximize expected utility and hence aims to submit an optimal ranking.

A voting method $F$ is \textit{manipulable at $\mathbf{U}$ by $i$} if there is some profitable manipulation of $F$ at $\mathbf{U}$ by $i$; and $F$ is \textit{manipulable} if there is some utility profile $\mathbf{U}$ and voter $i$ such that $F$ is manipulable at $\mathbf{U}$ by $i$. This notion of manipulability of $F$ coincides with the notion of manipulability of $F$ in \citealt{gibbard1977manipulation}  when we regard $F$ as a probabilistic voting method that assigns to each profile $\mathbf{P}$ the lottery $F_\ell(\mathbf{P})$.

\subsection{Voting Methods}\label{VMs}

In this paper, we focus on eight preferential voting methods:

\textbf{Plurality}: the winners are those candidates who receive the most first-place rankings from voters.

\textbf{Instant Runoff} with parallel-universe tiebreaking (IRV-PUT): if more than half of the voters rank the same candidate $a$ in first place, then $a$ wins; otherwise a candidate $a$ is an IRV-PUT winner if for one of the candidates $b$ who received the fewest first-place votes in $\mathbf{P}$, $a$ is the IRV-PUT winner in the profile $\mathbf{P}_{-b}$.

\textbf{Borda}: a candidate receives $0$ points from each voter who ranks them in last place, $1$ point from each voter who ranks them in second-to-last place, 2 points from each voter who ranks them in third-to-last place, etc., yielding a \textit{Borda score}; the candidates with maximal Borda score win.

\textbf{Black's}: if there is a Condorcet winner, that candidate wins; otherwise the Borda winners win.

\textbf{Minimax}: the winners  are those candidates $a$ who minimize the quantity  $\mathrm{max}\{\mathrm{Margin}_\mathbf{P}(b,a)\mid b\in X\}$.

\textbf{Nanson}: iteratively eliminate all candidates with less than average Borda score until there are no such candidates. The remaining candidates are Nanson winners. 

\textbf{Split Cycle}: the \textit{margin graph} of a profile is the weighted directed graph whose nodes are candidates with an edge from $a$ to $b$ of weight $k$ if $a$ has a positive margin of $k$ vs.~$b$. In each cycle in the graph (simultaneously), delete the edges with minimal weight. Then the candidates with no incoming edges are the winners.

\textbf{Stable Voting}: if there is only one Split Cycle winner in $\mathbf{P}$, they win; otherwise find the pairs of candidates $(a,b)$ where $a$ is a Split Cycle winner with the maximal margin of $a$ vs.~$b$ such that $a$ is a Stable Voting winner in  $\mathbf{P}_{-b}$, and declare $a$ a  winner in~$\mathbf{P}$.

Plurality, Instant Runoff, and Borda are perhaps the most famous of preferential voting methods. Plurality has been used for many centuries, and Instant Runoff and Borda date back to at least the 18th century. For Instant Runoff, there are multiple ways of handling ties in the number of first-place votes.\footnote{\label{SimulElim}Another version of IRV, as in \citealt{Taylor2008}, simultaneously eliminates \textit{all} candidates with the fewest first-place votes in a given round, unless all candidates have the same number of first-place votes, in which case all win.} The PUT version is popular in computational social choice \cite{Wangetal2019}. 

All of the other methods are \textit{Condorcet consistent} in the sense that if there is a Condorcet winner, that candidate is the unique winner according to the method. Plurality, Borda, and IRV-PUT all violate Condorcet consistency. Black's \cite{black1958theory} and Minimax \cite{Simpson1969,Kramer1977} are two of the most well known Condorcet methods.

The Nanson\footnote{There are two versions of  Nanson, one that removes all candidates with below average Borda score (Strict Nanson) and one that removes all candidates with less than or equal to average Borda score (Weak Nanson) unless doing so eliminates everyone. Apparently Nanson himself had in mind Weak Nanson \cite{Niou1987}, but most of the literature in computational social choice focuses on Strict Nanson (see \citealt{Brandt2016}), so we do as well.} method \cite{Nanson1882} has previously been studied in connection with strategic voting. In \citealt{narodytska2011manipulation}, it is shown that the problem of manipulating Nanson (and the related Baldwin method) so as to elect a desired candidate is NP-hard when the number of candidates is allowed to increase.

 Finally, we include the recently proposed Split Cycle voting method \cite{HP2023b}, whose manipulability has been studied in \citealt{durand2023coalitional}, as well as one of its refinements,\footnote{A voting method $F$ refines a method $G$ if for any profile $\mathbf{P}$, $F(\mathbf{P})\subseteq G(\mathbf{P})$. Other refinements of Split Cycle include Beat Path \cite{Schulze2011} and Ranked Pairs \cite{Tideman1987}.} the Stable Voting method \cite{HP2023}. These methods satisfy not only Condorcet consistency but also the stronger property---violated by Minimax---of Smith consistency, meaning that their winners always belong to the Smith set, the smallest set of candidates such that every candidate inside the set beats every candidate outside the set head-to-head. No previous work has studied the manipulability of Stable Voting, so studying this method tests if our approach can predict the manipulability of a method as measured in other ways in the future.

\section{Learning to Manipulate}

How difficult is it for a computationally bounded agent to learn to manipulate against a given voting method under limited information? In this paper, we study this question through training and evaluating many multi-layer perceptrons (MLPs) with increasing numbers of learnable parameters. These MLPs act as function approximators for profitable manipulation policies for a given voting method and type of limited information. We can evaluate the manipulation resistance of a voting method by the size and complexity of the MLP required to learn a profitable manipulation policy, as well as the average profitability of learned policies.

We use MLPs as our underlying architecture because they impose no structural assumptions on the input data. This flexibility is particularly important in our experiments, as we compare a variety of types of limited information, each with its own structure (see below). To ensure that our results in this initial investigation are not biased toward any specific type of information, we chose an architecture that does not depend on structural assumptions about sequence and order (as with transformers) or local patterns (as with CNNs).

\subsection{Implementation Details}

We optimize weights $\theta$ of an MLP $f$ whose input $x$ consists of a utility function $\mathbf{U}_v$  for our manipulating voter $v$, as well as some limited information $I$ about the full utility profile $\mathbf{U}$ (see below). We apply a softmax to the output of the MLP to generate a probability distribution over all possible actions of $v$, namely the $m!$ possible rankings, labeled 0 through $m!-1$, that  $v$ can submit, as in (\ref{MLPoutput}):
\begin{equation}
f_\theta (x) = \pi(x) = \left[ \mathbb{P} ( 0 | x), 
\mathbb{P} ( 1 | x), \dots, \mathbb{P} ( m! - 1 | x)
\right].
\label{MLPoutput}
\end{equation}
Note that $v$ is allowed (though there is no special incentive) to submit the sincere ranking given by their utility function.

\subsubsection{Probability models for profiles}

To generate utility profiles for our experiments described below, we first used a standard \textbf{uniform utility model} (see, e.g., \citealt[p.~16]{Merrill1988}): for each voter independently, the utility of each candidate for that voter is drawn independently from the uniform distribution on the $[0,1]$ interval. 

We also used a \textbf{spatial 2D model}: each candidate and each voter is independently placed in $\mathbb{R}^2$ according to the multivariate normal distribution  (as in~\citealt{Merrill1988}) with no correlation between the two dimensions; the utility of a candidate for a voter is the negative of the square of the Euclidean distance between the candidate and the voter (using the \textit{quadratic proximity} utility function as in \citealt[p.~21]{Merrill1999}). 

Finally, we used the \textbf{normalized Mallows model} from \citealt[\S~2.2]{Boehmer2023} with the dispersion parameter set to $\phi=.8$. This Mallows model generates preference profiles, which we turn into utility profiles by randomly choosing for each ballot (though we really only need to do this for the manipulator $v$) a utility function that induces the ballot as in (\ref{InducePref}). We do so by choosing $m$
 utilities from the uniform distribution on $[0, 1]$ and then assigning the highest utility to the top ranked candidate on the ballot, the second highest utility to the second ranked candidate, etc.

 These utility profiles can then be parameterized as 2D matrices, $U \in \mathbb{R}^{n \times m}$, where $n$ is the number of voters, $m$ is the number of candidates, and $U[i, c] \in \mathbb{R}$. To select the utility function for a given voter $i$ is to select the row $U_i$.

\subsubsection{Choices of Limited Information}\label{LimitedInfo}

We experimented with providing different types of input to the MLP, including the following types that are often taken to be natural forms of polling information in voting theory (see, e.g.,  \citealt[\S~2.2]{reijngoud2012voter}, \citealt{Veselova2023}):

\begin{itemize}
\item the \textbf{plurality score} of each candidate $a$, defined as the number of voters whose favorite candidate is $a$. $I \in \mathbb{N}^{m}$
\item the \textbf{plurality ranking}, i.e., the ordinal ranking of the candidates by their plurality scores. $I \in \mathbb{N}^{m}$
\item the \textbf{margin matrix} of dimension $m\times m$, where an integer $k$ in the $(a,b)$-coordinate of the matrix indicates that the margin of $a$ vs. $b$ is $k$. $I \in \mathbb{Z}^{m \times m}$
\item the \textbf{majority matrix}, obtained from the margin matrix by replacing all positive entries by $1$ and all negative entries by $-1$. $I \in \{-1,0, 1\}^{m \times m}$
\item the \textbf{qualitative margin matrix}, obtained from the margin matrix by replacing each positive margin by its rank in the ordering of margins from smallest to largest, and then adding negative entries so that the resulting matrix is skew-symmetric. $I \in \mathbb{Z}^{m \times m}$
\item the \textbf{sincere winners}, i.e., the candidates who would win according to the sincere profile $\mathbf{P}$. $I \in \{0, 1\}^{m}$
\end{itemize}

\noindent These additional inputs are flattened and concatenated to $v$'s utility function before being used as input to the MLP.

Note, crucially, that the full preference profile $\mathbf{P}$ is not uniquely determined by any of the types of limited information above. If the manipulating voter $v$ had full knowledge of $\mathbf{P}$, they could simply compute which of the $m!$ linear orders would be optimal to submit given $\mathbf{P}_{-i}$. But if $v$ has only some limited information $I$, e.g., the margin matrix, then there is a \textit{set} $\mathbb{P}(I)$ of preference profiles that could have generated the given limited information~$I$. Where $\mu$ is the probability distribution on preference profiles induced by our probability distribution on utility profiles, it might not be feasible for $v$ to sample sufficiently many profiles from the conditional distribution $\mu(\cdot\mid \mathbb{P}(I))$ in order to obtain a good estimate of the expected utility of submitting a linear order, when the limited information is $I$, for each of the $m!$ linear orders.\footnote{For some types of limited information, such as the \textbf{plurality ranking}, a simple rejection sampling approach is feasible: sample profiles according to $\mu$ and throw out those that do not realize the given plurality ranking. Since there are relatively few plurality rankings, one rapidly acquires many profiles realizing a given plurality ranking. By contrast, since there are vastly more margin matrices, it may be necessary to sample a huge number of profiles before finding sufficiently many that realize a given margin matrix.} Rather than investigating such a sampling approach, in this paper we will train an MLP to input $I$ and output an optimal ranking given $v$'s utilities.

Also note that information sufficient for determining the sincere winner according to a voting method---e.g., the plurality ranking for Plurality, the qualitative margin matrix for Minimax, Split Cycle, and Stable Voting, etc.---is not necessarily  sufficient for determining \textit{who would win} after a particular manipulation.

\subsubsection{Labeling}

We framed the learning objective as a classification task. Given a voting method and utility profile, we used the following labeling of each of the $m!$ possible rankings $v$ could submit:
\begin{itemize}
\item \textbf{optimizing labeling}: all optimal rankings to submit are labeled by $1$, and all other rankings are labeled by $0$.\footnote{We duplicated our experiments with a \textbf{satisficing labeling} (if there are profitable manipulations, they are labeled by $1$, and all other rankings are labeled by~$0$; otherwise all rankings that do at least as well as the sincere ranking are labeled by $1$ and all others by $0$), but  MLPs trained with this labeling had qualitatively indistinguishable results from those trained with the optimizing labeling.}
\end{itemize}

The output of our MLPs is a distribution over all $m!$ rankings given some information $x$ about the current utility profile. It is equally valid for our agent to choose any of the positively-labeled rankings. We treat the binary labelings as a mask over the rankings and reduce the distribution $\pi (x)$ to two values: the probability of choosing a positively-labeled ranking or not. We compute the final loss as the mean-squared error between this reduced distribution and the distribution assigning probability 1 to choosing a positively-labeled ranking and 0 to choosing a non-positively-labeled~ranking.\footnote{After a first run of our experiments, we reran the entire experiment for the uniform utility model with the final loss computed as the binary cross-entropy loss, but this did not lead to any substantial difference in the performance of trained MLPs that would affect our conclusions (see the GitHub repository for details).}

\subsection{Evaluation}

To evaluate how well a given MLP has learned to manipulate, we must convert its output distribution over rankings into a single ranking. To do so, we use the following decision rule:
\begin{itemize}
    \item \textbf{argmax}: select the ranking with the maximum probability in the output of the MLP.
\end{itemize}

As our metric for the profitability of the MLP's decision, we use the difference between the left and right-hand sides of (\ref{ManipEq}) normalized by the greatest possible utility difference according to~$\mathbf{U}_i$:
\begin{equation}\frac{\mathbf{EU}_i (F_\ell(\mathbf{P}'_i, \mathbf{P}_{-i}))-\mathbf{EU}_i (F_\ell(\mathbf{P}))}{ \mathrm{max}(\{\mathbf{U}_i(x)\mid x\in X\}) - \mathrm{min}(\{\mathbf{U}_i(x)\mid x\in X\})}.\label{ProfitabilityEq}\end{equation}
We call this the \textit{profitability of $\mathbf{P}'_i$ with respect to $F,\mathbf{U},i$}. 

The normalization in (\ref{ProfitabilityEq}) is the standard normalization for \textit{relative utilitarianism} \cite{DhillonMertens1999}, which we use to compare utility differences across profiles. Indeed, (\ref{ProfitabilityEq}) is equivalent to taking the difference in the \textit{Kaplan-normalized} \cite[p.~470]{d'Aspremont2002} expected utilities of submitting $\mathbf{P}'$ and of submitting $\mathbf{P}$, where $\hat{u}=\mathrm{max}(\{\mathbf{U}_i(x)\mid x\in X\})$, $\check{u}=\mathrm{min}(\{\mathbf{U}_i(x)\mid x\in X\})$:

\[\frac{\mathbf{EU}_i (F_\ell(\mathbf{P}'_i, \mathbf{P}_{-i}))-\check{u}}{ \hat{u} -\check{u}} - \frac{\mathbf{EU}_i (F_\ell(\mathbf{P}))-\check{u}}{\hat{u} - \check{u}}.\]

Note that for a particular decision by an MLP, the numerator of (\ref{ProfitabilityEq}) may be \textit{negative}, i.e., the MLP may be worse off by submitting an insincere ranking $\mathbf{P}'$ than they would have been by submitting the sincere ranking $\mathbf{P}$.

For a given trained MLP, we sample utility profiles according to one of our probability models and compute the average profitability of the MLP's submitted rankings. We take the average profitability that is achieved against a given voting method as a measure of a voter's incentive to manipulate against that voting method. This is a more revealing measure than the frequency with which the trained MLP manipulates against a voting method, since many of the MLP's manipulations may be no better than sincere~voting.

For the number of profiles to sample, we continued sampling  until the estimated standard error of the mean \cite[\S~3.2]{RobertCasella2004} for profitability fell below 5e-4, resulting in small error bars (see Figure \ref{UniformPlurScoreMaj}).\footnote{The number of profiles sampled for evaluation of an MLP was at least 4,096 with an average of approximately 36,000 and a standard deviation of approximately 31,000.}

\subsubsection{Baselines}

For baseline comparisons, we consider an agent with full information about a profile and unbounded computational resources, who always picks one of the optimal rankings to submit.  We estimated the average profitability as in (\ref{ProfitabilityEq}) of this agent's submitted ranking across many sampled elections, where as above we continued sampling until the estimated standard error of the mean fell below {5e-4}.

\subsection{Training Setup}

For each voting method $F$, each $n\in\{5,6,10,11,20,21\}$, each  $m\in \{3,4,5,6\}$, each choice of an input type for the MLP, and each choice of a model size (see the $x$-axis of  Figure \ref{UniformPlurScoreMaj}), we trained one or more ``generations'' of MLPs with that model size to manipulate elections with $n$ voters and $m$ candidates run using $F$, resulting in over 100,000 trained MLPs. For a given generation, we used the same initialization of MLP weights and the same training, validation, and evaluation profiles for every MLP for $n$ voters and $m$ candidates. Across generations, we varied the initialization of MLP weights and used different training, validation, and evaluation profiles, to provide reassurance that our results were not due to lucky initial weights or profiles. All elections and labels were pre-computed so training could rely fully on the GPU.

We experimented with training the MLPs using different numbers of iterations---between 100 and 1000---different learning rates---1e-3, 3e-3, and 6e-3---and different batch sizes---256 and 512. We hand-tuned these hyper-parameters to try to maximize the performance of all MLPs across all voting methods and profile sizes. For the final training run reported here, we use a batch size of 512 and a learning rate of 6e-3. We train all models for at least 220 iterations and then terminate training with an early stopping rule: after every 20 iterations, we measure the average profitability on a validation batch of 4,096 elections. If 10 validation steps pass without an improvement of at least .001 in average profitability of the submitted ranking, we terminate training.\footnote{The average number of training iterations of each MLP was approximately 600 with a standard deviation of approximately 250.} 

\subsubsection{Computing Infrastructure}

All code was written in Python using PyTorch, version 2.0.1, and the pref\_voting library (pypi.org/project/pref-voting/), version 0.4.42 or later. Training and evaluation were parallelized across nine local Apple computers with Apple silicon, the most powerful equipped with an M2 Ultra with 24-core CPU, 76-core GPU, and 128GB of unified memory, running macOS 13, as well as up to sixteen cloud instances with Nvidia A6000 or A10 GPUs running Linux Ubuntu 18.04.

\section{Results}

\begin{figure*}[h!]

\begin{center} Uniform Utility Model\end{center}
\includegraphics[scale=.29]{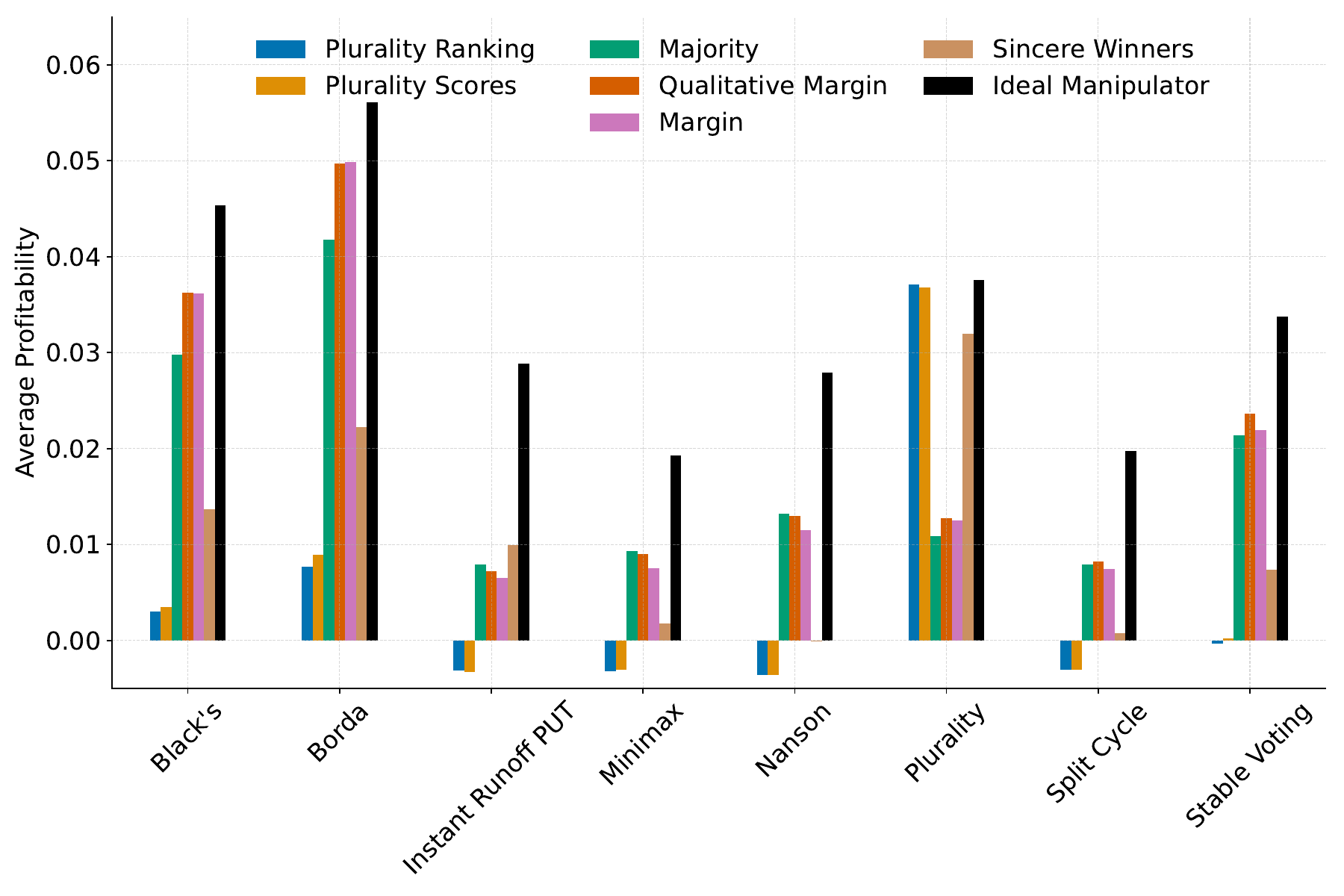}
\includegraphics[scale=.29]{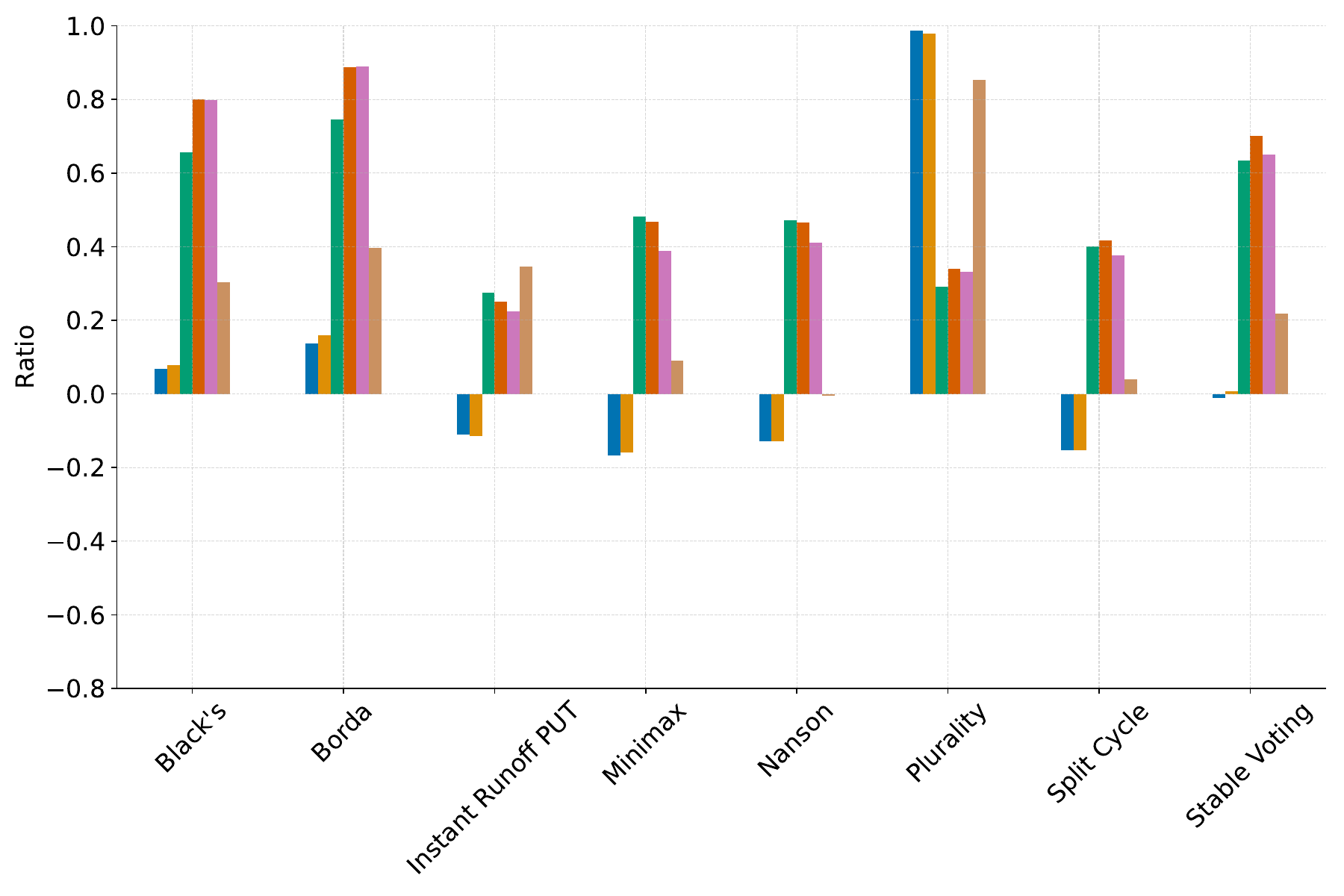}
\begin{center} Mallows Model\end{center}
\includegraphics[scale=.29]{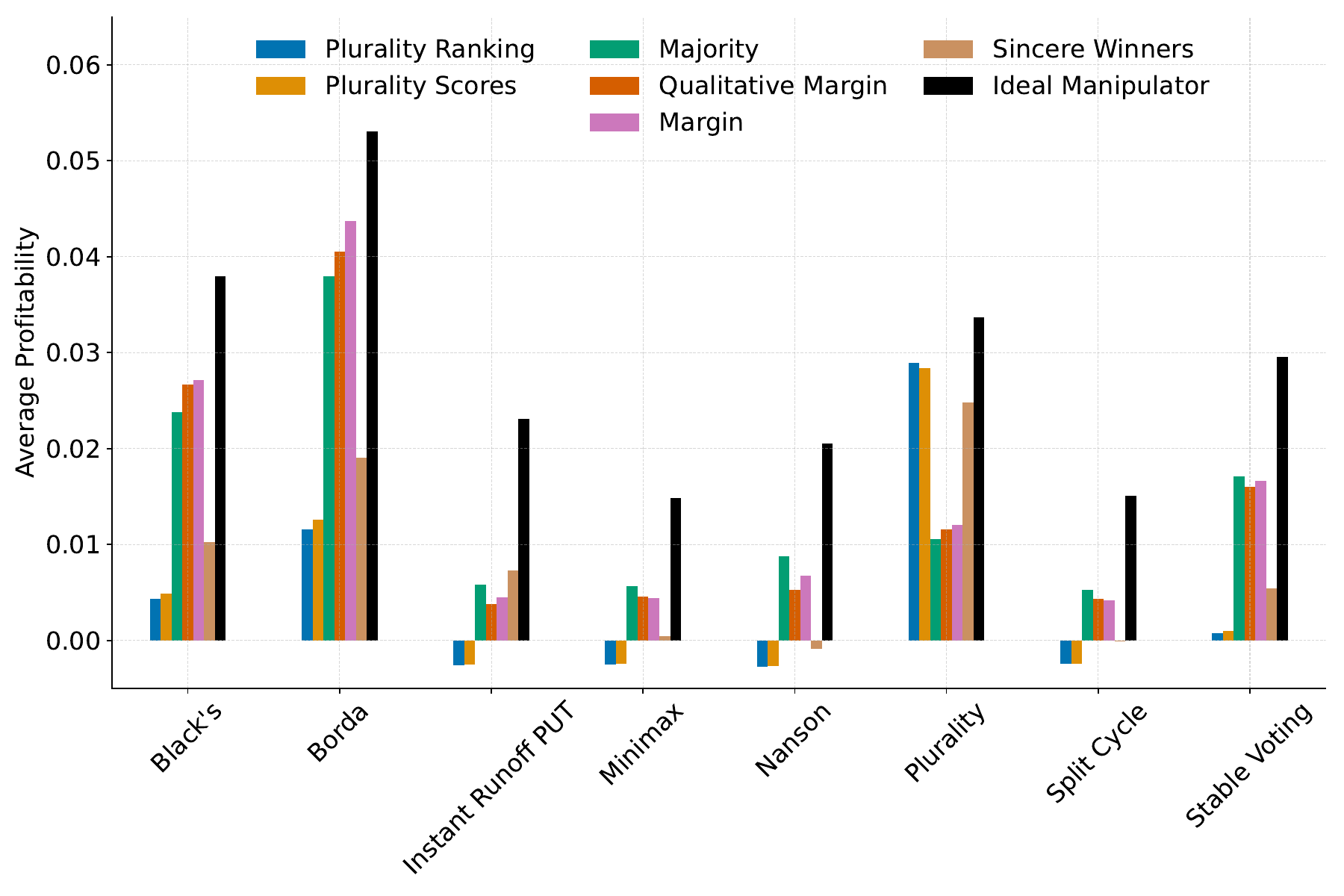}
\includegraphics[scale=.29]{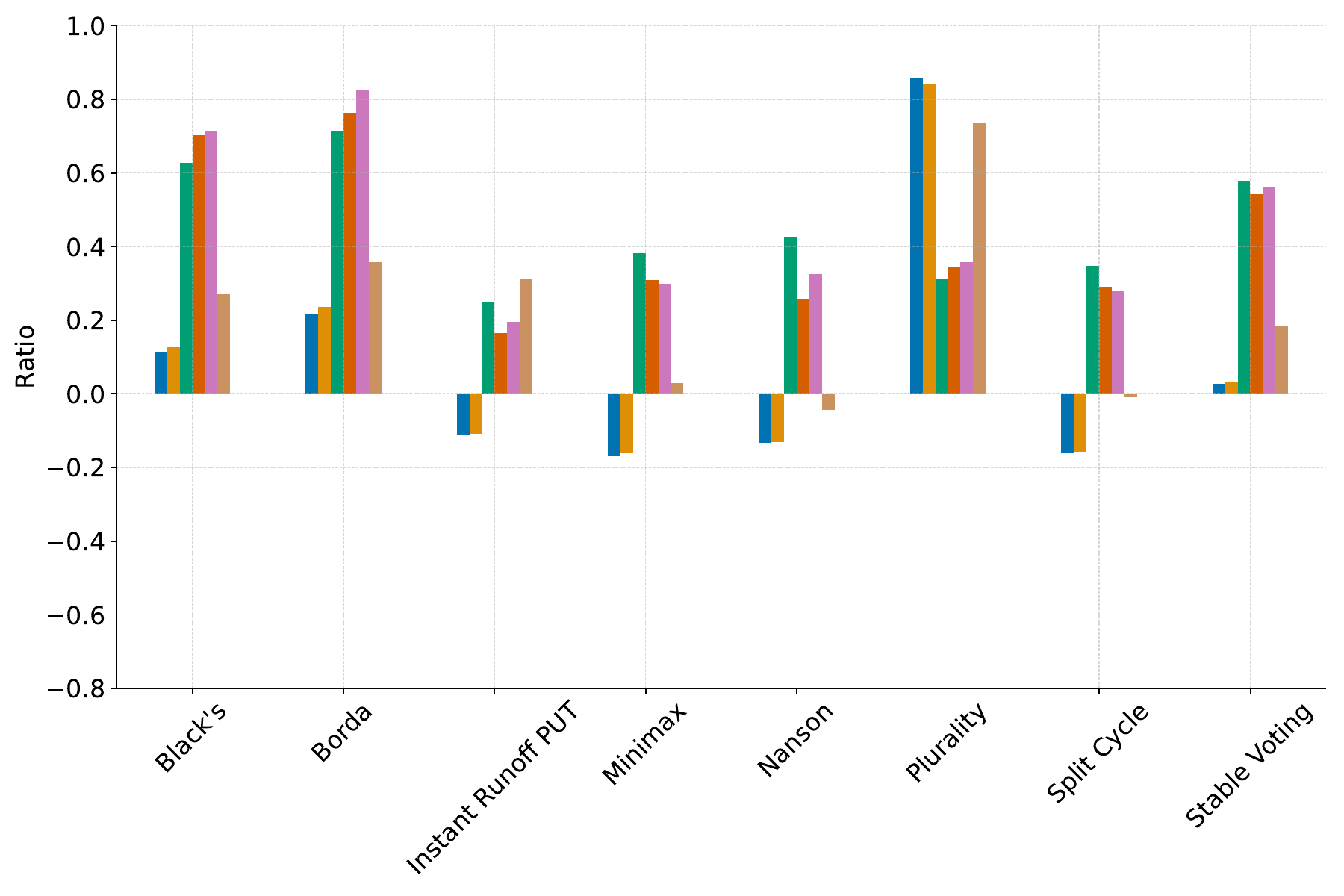}
\begin{center} Spatial 2D Model\end{center}
\includegraphics[scale=.29]{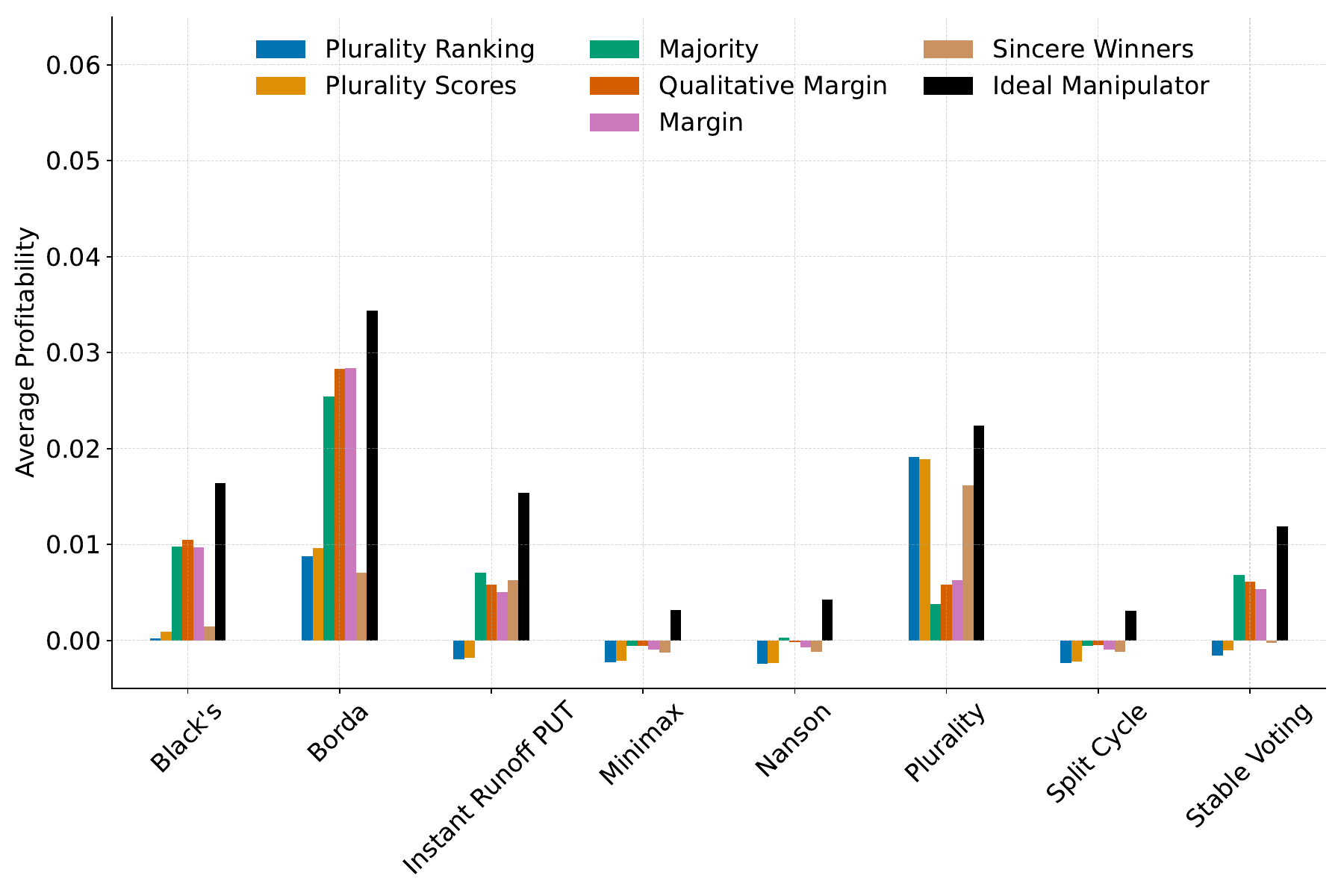}
\includegraphics[scale=.29]{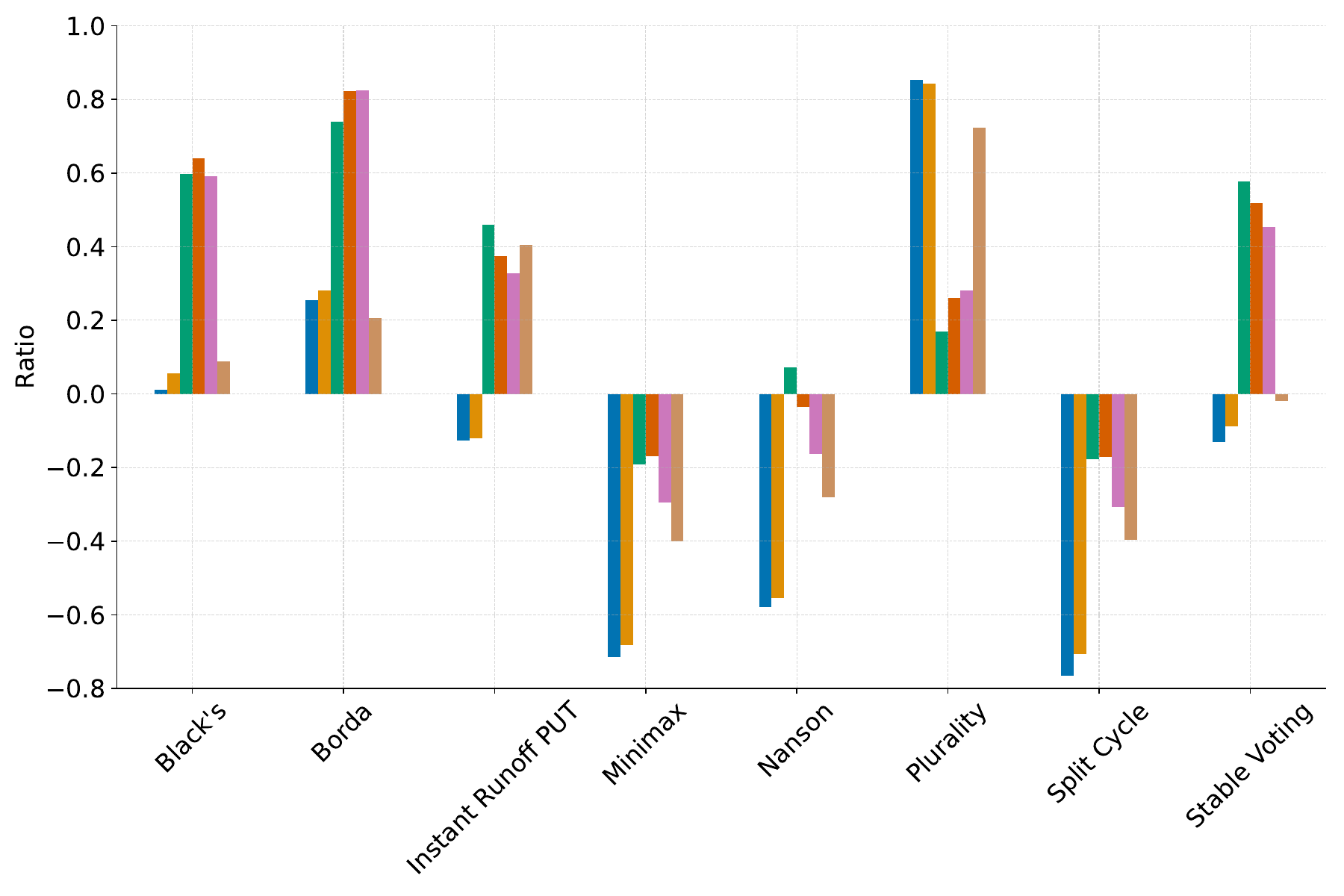}

\caption{Left: the average profitability of submitted rankings by the best performing MLP with any hidden layer configuration for a given voting method and information type, averaging over 3--6 candidates and 5, 6, 10, 11, 20, and 21 voters. Right: the ratio of the average profitability of the MLP's submitted ranking to that of the ideal manipulator's submitted ranking.}\label{AgentInfosFig}
\end{figure*}

The average profitability of submitting the optimal ranking in each election  (see Baselines above) with different voting methods is shown by the black bars in Figure \ref{AgentInfosFig}. The other colored bars are for MLP-based manipulators with different types of limited information. Figure~\ref{AgentInfosFig} shows data for the performance (averaging over different numbers of candidates and voters) of the \textit{best performing} MLPs (for each number of candidates, number of voters, voting method, and choice of information) with any hidden layer configuration.  

\begin{figure*}
\centering 
\includegraphics[scale=0.325]{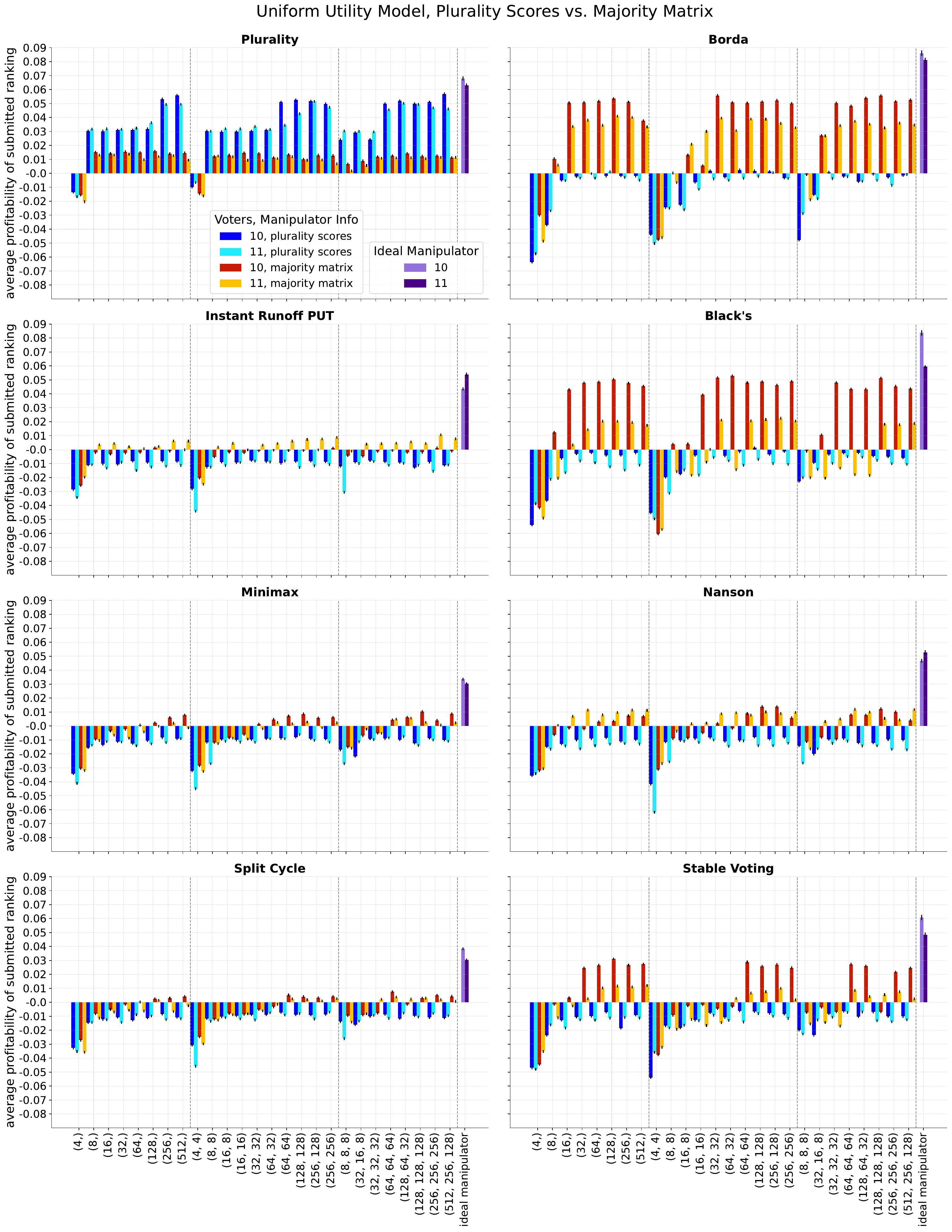}
\caption{Results using the \textbf{uniform utility model} with 6 candidates and 10/11 voters for MLPs manipulating on the basis of the \textbf{plurality scores} or \textbf{majority matrix}. Error bars indicate twice the estimated standard error of the mean. Hidden layer configurations of trained MLPs are shown on the x-axis. Versions of this figure for the Mallows model,  spatial 2D model, and different types of limited information appear in Supplementary Figures A.2--B.3.}\label{UniformPlurScoreMaj}
\end{figure*}

Figure \ref{UniformPlurScoreMaj} and Supplementary Figures~A.2--B.3 show the performance of MLPs with each of the 26 different hidden layer configurations we tested, focusing on 6 candidates, 10/11 voters, and different choices of limited information. All Supplementary Figures cited are available at {\small github.com/epacuit/ltm/blob/main/supplementary-figures.pdf}.

These figures all cover the first generation of trained MLPs. Results for the second and third generations of trained MLPs are qualitatively similar to the first (see the GitHub repository), though we lack a sufficient number of generations to make quantitative statistical claims.\footnote{However, we can make claims of statistical significance about the differences in performance between two particular, trained MLPs, e.g., one trained using the \textbf{majority matrix} vs. one trained using the \textbf{plurality scores} (see the GitHub repository).}  The following qualitative highlights are robust across generations. All claims are implicitly qualified to apply to elections with 3--6 candidates and 5-21 voters.

\newpage $\,$ \newpage $\,$ \newpage

\subsection{Differences across probability models}\label{SpatialSection}

Comparing the results for the three probability models in Figure \ref{AgentInfosFig}, we see that MLPs achieved the highest profitability in manipulating against voting methods in elections generated by the \textbf{uniform utility model}, followed by the \textbf{Mallows model}, followed by the \textbf{spatial 2D model}. While the graphs for Mallows are roughly scaled down versions of those for the uniform model, it is striking that for the spatial 2D model,  even the best MLPs could not learn to profitably manipulate against Minimax, Nanson, and Split Cycle.\footnote{A natural thought to explain this is that Minimax, Nanson, and Split Cycle are Condorcet consistent, and there is a high frequency of Condorcet winners under the spatial 2D model (yet this must be squared with the results for Stable Voting). However, even when there is a Condorcet winner in $\mathbf{P}$ and we are using a Condorcet voting method, a voter may still have an incentive to submit an insincere ranking in order to create a majority cycle, possibly resulting in a different winner. On the other hand, such possibilities are evidently rare and difficult to learn to exploit.} On the other hand, \textit{the comparative usefulness, for manipulating against each voting method, of the different types of limited information is largely the same} under all three models (this is true even for Minimax, Nanson, and Split Cycle under the spatial model, looking at which types of information produce less negative results). We conjecture that these findings about types of limited information are robust across other standard probability models as~well.

\subsection{The importance of majorities} With the \textbf{uniform utility} and \textbf{Mallows} models, sufficiently large MLPs learned to profitably manipulate \textit{all} eight voting methods on the basis of knowing only the \textbf{majority matrix}, though the profitability of such manipulation varied dramatically across methods. Interestingly, we did not find a substantial increase in profitability of manipulation for MLPs that learned to manipulate based on the more informative \textbf{margin matrix} instead of the \textbf{majority matrix}, except in the case of Borda and Black's (especially for 6 candidates, as shown in Supplementary Figures B.1--B.3). In fact, the \textbf{qualitative margin matrix} was about as useful as the \textbf{margin matrix} for learning to manipulate Borda and Black's.

\subsection{The limited usefulness of plurality scores} While knowing the plurality scores is obviously useful for manipulating Plurality and somewhat useful for several methods for 3 candidates (though less useful for 4 or 5), it was insufficient in 6-candidate elections for profitably manipulating methods other than Plurality (though Borda may be barely manipulable in this~case), as shown in  Figure \ref{UniformPlurScoreMaj} and Supplementary Figures A.2--A.3. Moreover, in the case of manipulating Plurality, learning to manipulate on the basis of the \textbf{plurality ranking} led to profitability comparable to learning on the basis of the plurality scores themselves (see Figure~\ref{AgentInfosFig}).

\subsection{Highly manipulable vs. resistant methods} Plurality and especially Borda have long been regarded as highly manipulable.\footnote{Borda declared that his method was ``intended for only honest men'' \cite{black1958theory}. Based on heuristic algorithms for manipulating Borda,  Walsh \citeyearpar{walsh2011computational} concludes that  ``Borda voting can usually be manipulated with relative ease'' (p.~13). For empirical results on manipulation of Borda by humans, see \citealt{kube2009voters}.} Our results show that this is so even under limited information, e.g., the majority matrix. The manipulability of Borda seems to infect Black's method as well, as it uses Borda when there is no Condorcet winner.

IRV-PUT was quite resistant to manipulation on the basis of limited information, despite the fact that it is more manipulable than some others by an ideal manipulator. In addition, Minimax and  Split Cycle  stood out for their resistance to manipulation, especially under the spatial 2D model. For the uniform utility and Mallows models, it is noteworthy that while Minimax and Split Cycle were not much more profitably manipulable than IRV-PUT on the basis of the \textbf{majority matrix}, \textbf{qualitative margin matrix}, or \textbf{margin matrix}  in absolute terms, MLPs came closer to the ideal manipulator for learning to manipulate Minimax and Split Cycle based on this information than for IRV-PUT (see Figure~\ref{AgentInfosFig}), which is more manipulable by an ideal manipulator. Another noteworthy difference is that an MLP with only the \textbf{sincere winners} information can achieve between 30-40\% of the profitability of an ideal manipulator when manipulating against IRV-PUT (see Figure~\ref{AgentInfosFig}), whereas the average profitability of manipulation by an MLP with sincere winners information is negative or barely positive when manipulating against Minimax or Split Cycle.

 \subsection{The subtleties of tiebreaking} As noted in Footnote \ref{SimulElim}, there are different ways of dealing with ties in first-place votes for Instant Runoff. This actually leads to significant differences with respect to single-voter manipulability under 3 candidates and an even number of voters (see Supplementary Figure C.1). Of course, manipulation by a single voter is only possible in very close elections, in which case ties matter. 

\subsection{Parity of the number of voters} The parity of the number of voters is a key factor for some  methods. This is most striking for  Stable Voting (also see Black's), which is barely manipulable with 11 voters but more manipulable with 10 (see Figure \ref{UniformPlurScoreMaj} and Supplementary Figures A.2--A.3). A key difference is that with 10 voters, it is possible to have margins of \textit{zero} between candidates, in which case a single voter has more manipulative power under Stable Voting, which produces fewer tied elections in the presence of zero margins than other methods like Minimax.  

\begin{figure}
\begin{center}
\includegraphics[scale=0.3025]{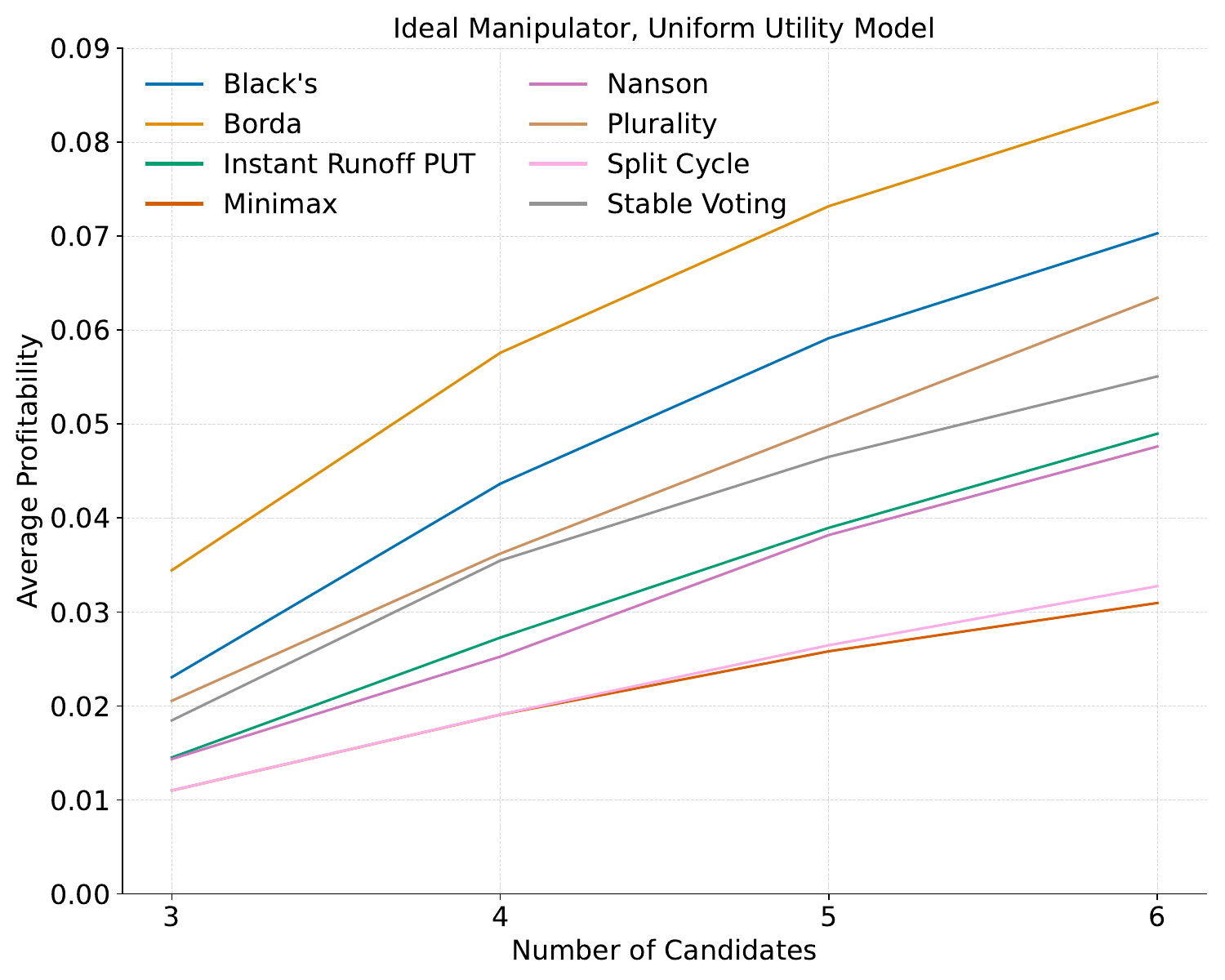}\vspace{.1in}
\includegraphics[scale=0.3025]{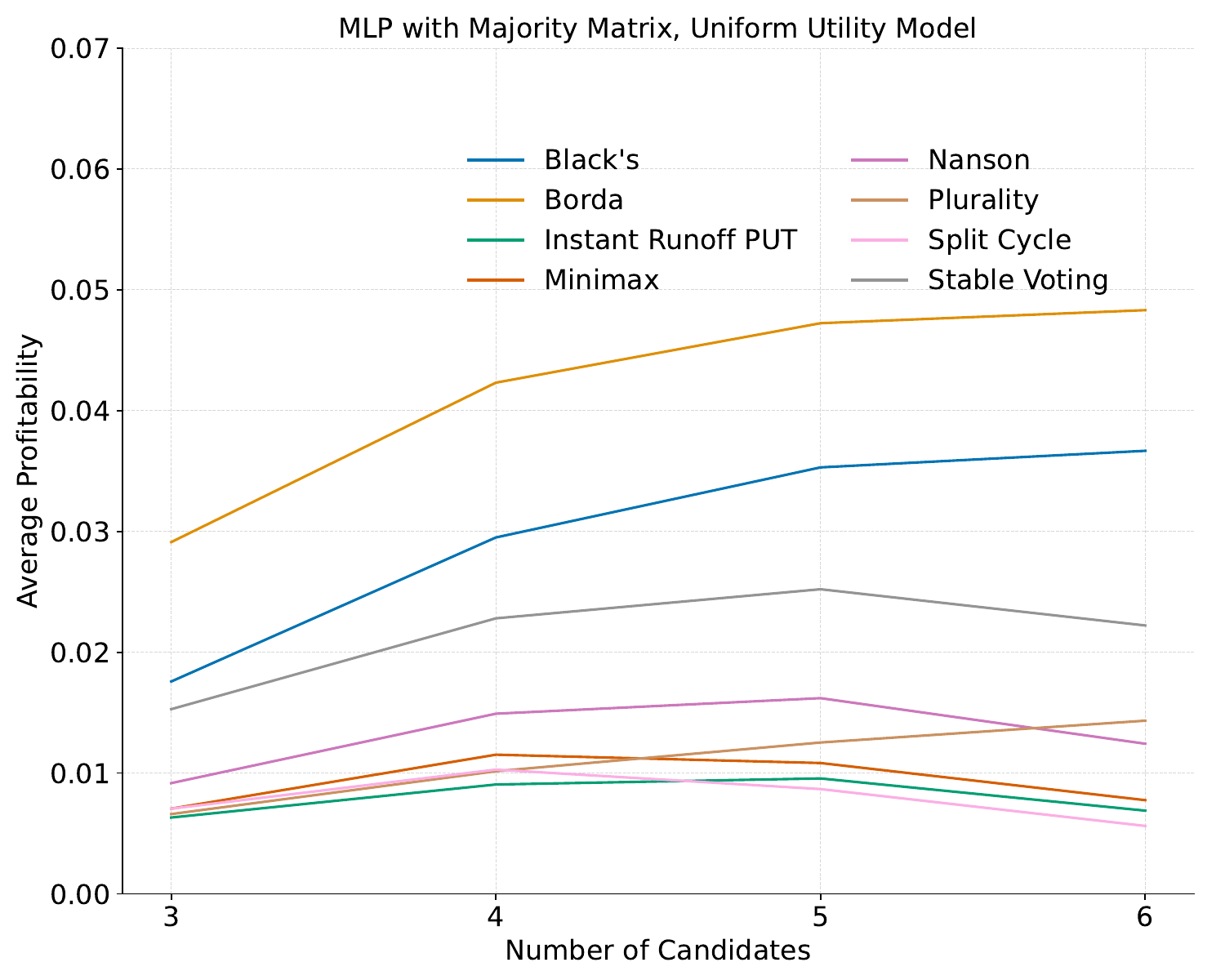}\vspace{.1in}
\includegraphics[scale=0.3025]{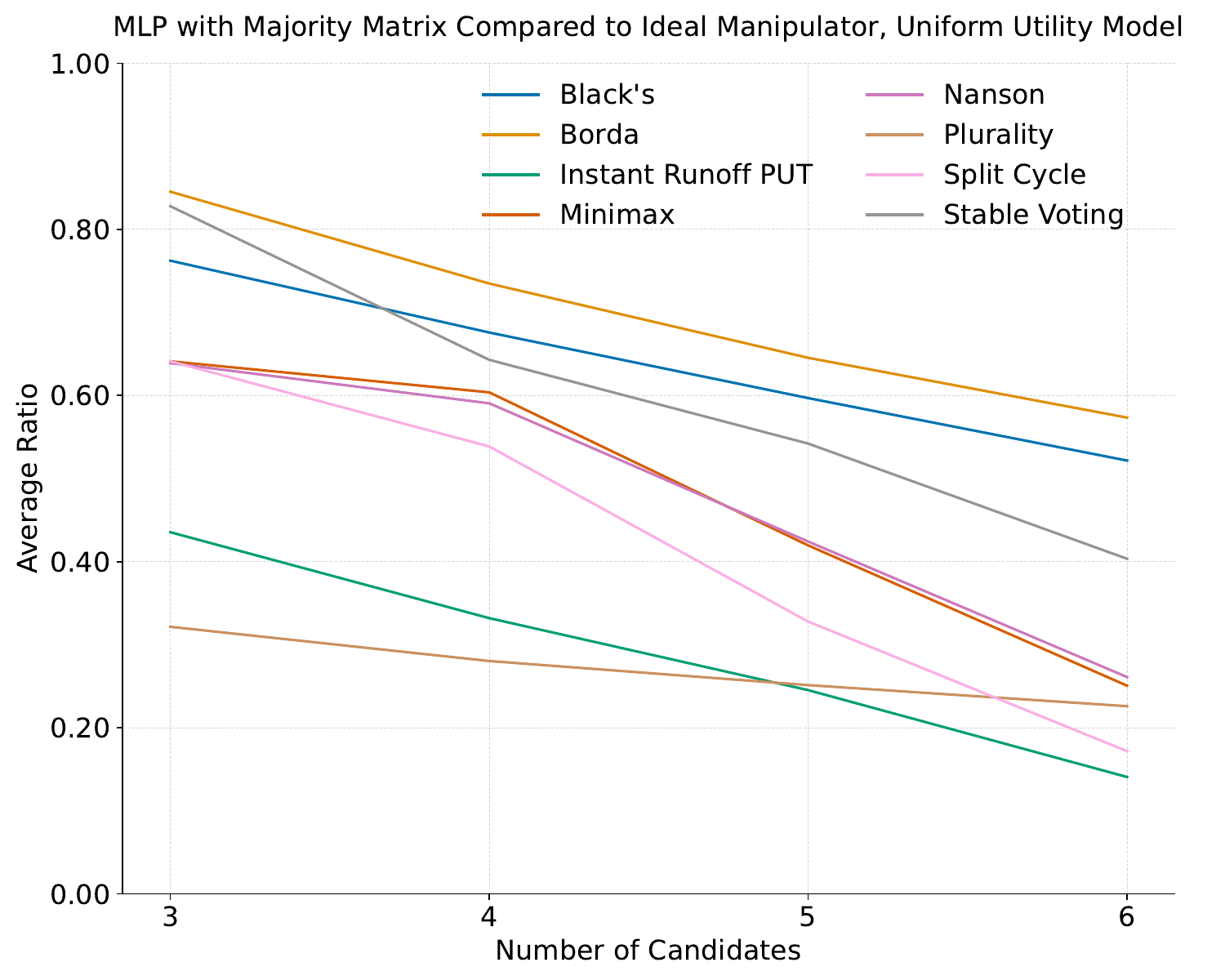}
\end{center}

\caption{Top: average profitability of submitted rankings by an ideal manipulator. Middle: average profitability by the best performing MLP with any hidden layer configuration using the \textbf{majority matrix} information, averaging over 5, 6, 10, 11, 20, and 21 voters. Bottom: the ratio of the average profitability of the MLP's submitted ranking to the average profitability of the ideal manipulator's submitted ranking. Versions of this figure for the Mallows model and the spatial 2D model appear in Supplementary Figures D.2--3.}\label{NumCandsUniform}
\end{figure}

\subsection{Effect of the number of candidates} For an ideal manipulator, manipulation becomes more profitable as the number of candidates increases (see Figure \ref{NumCandsUniform} and Supplementary Figures D.2--D.3). The same can be true for an MLP-based manipulator up to a point, e.g., 4 or 5 candidates for most voting methods when using the majority, qualitative margin, or margin information. However, the ratio between the profitability of the MLP-based manipulator's submitted rankings and those of the ideal manipulator declines as the number of candidates increases from 3 to 6 (again see the figures just cited). This is intuitive, as manipulation should increase in complexity with more candidates.

\subsection{Profitability and ease of learnability} Not only is it more profitable to manipulate, say, Borda than Stable Voting, but it is also easier to learn---in the sense of requiring smaller MLPs: whether we consider one, two, or three hidden layers, a smaller network is needed to learn to profitably manipulate Borda compared to Stable Voting on the basis of the majority or margin matrix (see  Figure \ref{UniformPlurScoreMaj} and Supplementary Figures A.2--B.3). 

\subsection{Worst-case complexity vs. learnability} Despite the NP-hardness (when we allow the number of candidates to increase) of deciding if one can manipulate Nanson so as to elect a desired candidate, it is still possible (under the uniform utility and Mallows models) to learn to manipulate Nanson to achieve an increase in expected utility. In this connection, it would be interesting to study learning to manipulate with more candidates.

\section{Conclusion}

In committee-sized elections (5-21 voters), MLPs can learn to vote strategically on the basis of limited information, though the profitability of doing so varies significantly between different voting methods. This serves as a proof of concept for the study of machine learnability of manipulation under limited information. There are a number of natural extensions for future work, including manipulation by a coalition of voters, additional probability models for generating elections, and training an MLP to manipulate in elections with varying numbers of voters and candidates (rather than training different MLPs for each choice of a number of voters and number of candidates). Our code is already set up to handle these extensions, which only require more compute. However, further research is needed on other questions: What if all agents in the election strategize? What is the social cost or benefit of the learned manipulations? How do different neural network architectures affect results? Finally, one limitation of the classification approach in this paper is that it is infeasible to apply to more than 6 candidates. To overcome this limitation, we plan to develop a reinforcement learning approach to learning to manipulate.

\section*{Acknowledgments}

For helpful feedback, we thank Dominic Hughes, the anonymous reviewers for SCaLA and for AAAI, and the participants in the Values-Centered AI seminar at the University of Maryland in Fall 2023, the logic seminar at the University of Maryland in Spring 2024, the Interdisciplinary Workshop on Computational Social Choice at the Center for Human-Compatible Artificial Intelligence in Spring 2024, and New Directions in Social Choice at EC 2024.

\bibliography{manipulation}

\end{document}